\documentclass[11pt,a4paper]{article}

\usepackage[margin=1in]{geometry}
\usepackage[T1]{fontenc}
\usepackage[utf8]{inputenc}
\usepackage{graphicx}
\usepackage{booktabs}
\usepackage{array}
\usepackage{amsmath}
\usepackage{amssymb}
\usepackage{hyperref}
\usepackage{textcomp}
\usepackage{caption}
\usepackage{authblk}
\usepackage{xcolor}

\hypersetup{
  colorlinks=true,
  linkcolor=black,
  citecolor=black,
  urlcolor=blue!60!black,
  pdftitle={Quantisation Reshapes the Metacognitive Geometry of Language Models},
  pdfauthor={Jon-Paul Cacioli}
}

\title{\textbf{Quantisation Reshapes the Metacognitive\\Geometry of Language Models}}

\author{Jon-Paul Cacioli\\
\small Independent Researcher, Melbourne, Australia\\
\small ORCID: 0009-0000-7054-2014}

\date{April 2026}

\begin{document}

\maketitle

\begin{abstract}
\noindent
We report that model quantisation restructures domain-level metacognitive efficiency in LLMs rather than degrading it uniformly. Evaluating Llama-3-8B-Instruct on the same 3,000 questions at Q5\_K\_M and f16 precision, we find that M-ratio profiles across four knowledge domains are uncorrelated between formats (Spearman $\rho = 0.00$). Arts \& Literature moves from worst-monitored (M-ratio $= 0.606$ at Q5\_K\_M) to best-monitored (1.542 at f16). Geography moves from well-monitored (1.210) to under-monitored (0.798). However, Type-2 AUROC profiles are perfectly stable across formats ($\rho = 1.00$), localising the restructuring to the M-ratio normalisation rather than the underlying discrimination signal.

This finding emerged from a pre-registered attempt to improve metacognition through domain-conditional training. We prescribed confidence-amplification SFT for the diagnosed weak domain, with matched-budget agnostic and wrong-prescription controls. All four confirmatory hypotheses were null (10{,}000 bootstrap resamples, seed $= 42$). The training successfully reshaped confidence distributions, doubling the NLP gap in Science from 0.076 to 0.152, but did not improve meta-$d'$ because the diagnostic profile did not transfer across formats.

Any system relying on domain-level M-ratio profiles has an unexamined dependency on inference format. Systems using AUROC$_2$ are safer. We release all code, pre-registrations, and trial-level data.
\end{abstract}

\section{Introduction}

Cacioli (2026a) demonstrated that metacognitive efficiency varies across knowledge domains within a single LLM. Using Type-2 signal detection theory (Maniscalco \& Lau, 2012; Fleming \& Lau, 2014), that study computed M-ratio (meta-$d'/d'$) per domain for four models and found that the weakest domain differs across models. For Llama-3-8B-Instruct, Science \& Technology had the poorest metacognitive efficiency (M-ratio $= 0.788$, versus 0.962 to 1.198 for other domains).

This finding invites an intervention. If you know which domain is under-monitored, train specifically on that domain. We call this the prescribe-don't-average principle. Current approaches to LLM confidence calibration apply uniform interventions, including temperature scaling (Guo et al., 2017), SFT on calibration-targeted data (Steyvers et al., 2025), and calibration-aware training losses (Parikh et al., 2026). None uses domain-level diagnostic information to target the intervention.

The present study attempted exactly this. We trained domain-conditional SFT on the diagnosed weak domain (Science), with three controls: domain-agnostic SFT with the same data budget, wrong-prescription SFT targeting a strong domain (Geography), and a low-learning-rate variant. All conditions were pre-registered on OSF before training.

The intervention failed. All four confirmatory hypotheses were null. But the failure was informative. It was caused by a quantisation format mismatch between diagnosis and evaluation that revealed a previously undocumented property of model quantisation.

The M1 diagnostic profiles were computed using Q5\_K\_M quantisation. The M2 evaluation used f16 format because the adapted models could not be quantised to Q5\_K\_M. At f16, the baseline M-ratio profile is completely different from the Q5\_K\_M profile. The intervention was correctly prescribed for a measurement format it was not evaluated under.

A same-question comparison on 3,000 items confirmed that this is a quantisation effect, not a question-set artefact. M-ratio profiles are uncorrelated across formats ($\rho = 0.00$). But AUROC$_2$ profiles are perfectly stable ($\rho = 1.00$). The restructuring is in the normalisation, not the discrimination. This dissociation is the central finding of the paper.

\section{Background}

\subsection{Metacognitive efficiency in LLMs}

Type-2 signal detection theory (Maniscalco \& Lau, 2012) distinguishes Type-1 performance (can the model answer correctly?) from Type-2 sensitivity (does the model's confidence discriminate correct from incorrect responses?). Meta-$d'$ measures the theoretical $d'$ an ideal observer would need to produce the observed confidence-correctness contingency given the model's actual Type-1 performance. M-ratio $=$ meta-$d' / d'$ normalises for task difficulty. M-ratio $= 1$ indicates optimal metacognitive efficiency. Below 1 indicates under-monitoring. Above 1 indicates confidence carries more information than expected.

Cacioli (2026a) applied this framework across four knowledge domains for four LLMs on TriviaQA and found substantial between-domain variability within each model.

\subsection{The prescribe-don't-average framework}

The prescribe-don't-average framework proposes four steps. Diagnose domain-specific metacognitive profiles via Type-2 SDT. Identify the weakest domain. Prescribe targeted training for that domain only. Evaluate whether the targeted domain improves without degrading others.

\subsection{Quantisation and confidence}

Model quantisation reduces numerical precision to decrease memory and computational cost. Standard evaluations show small accuracy losses under quantisation (Dettmers et al., 2022; Frantar et al., 2023). Recent work has begun examining effects beyond accuracy.

Proskurina et al. (2024) showed that GPTQ 4-bit quantisation decreases confidence on true labels, increases calibration error, and disproportionately affects samples where the unquantised model was already uncertain. Singh et al. (2025) found that confidence distributions and Adaptive Calibration Error shift under quantisation with varying severity across model families. Liu et al. (2025) demonstrated that quantised models can be recalibrated post-hoc, suggesting the confidence distortion is systematic rather than random.

These studies establish that quantisation affects the level of confidence and calibration. None examines whether quantisation changes the structure of confidence across content domains. Specifically, none tests whether the relative ordering of metacognitive efficiency across knowledge domains is preserved. This is the gap the present study addresses.

\subsection{Format-dependent measurement}

The finding that metacognitive measurement depends on format has precedent. Fleming and Lau (2014) cautioned that M-ratio is sensitive to $d'$ magnitude, producing unstable estimates when $d'$ is low. Cacioli (2026d) showed that 12 of 20 frontier LLMs produce inconsistent validity profiles across retrospective and prospective probe formats. Cacioli (2026e) demonstrated that transformers reproduce magnitude geometry but not scalar noise profiles. The present finding extends this pattern to the inference format dimension.

\section{Method}

\subsection{Pre-registration and design}

Two pre-registrations were posted to OSF before any training. Pre-registration 1 specified the pilot design and decision gate. Pre-registration 2 specified the full confirmatory grid. The training method (SFT) was selected via a pre-registered pilot in which SFT outperformed DPO and CATTO on meta-$d'$.

Seven conditions were pre-registered (Table \ref{tab:conditions}).

\begin{table}[h]
\centering
\caption{Pre-registered experimental conditions}
\label{tab:conditions}
\small
\begin{tabular}{clp{8cm}}
\toprule
\# & Condition & Training data \\
\midrule
1 & No intervention (baseline) & None \\
2 & Domain-conditional SFT & Science questions, confidence amplification on correct trials \\
3 & Domain-agnostic SFT & All domains proportionally, matched pair count \\
4 & Wrong-prescription SFT & Geography questions, confidence amplification \\
5 & Prompt-based domain cue & System prompt only \\
6 & Per-domain temperature scaling & Post-hoc calibration \\
7 & Domain-conditional SFT (LR $=$ 5e-6) & Same as Condition 2 \\
\bottomrule
\end{tabular}
\end{table}

Conditions 2, 3, 4, and 7 each used 651 SFT pairs. LoRA rank 16, alpha 32, 3 epochs, effective batch size 16.

\subsection{Data}

All datasets use TriviaQA questions classified into four domains by Llama-3-8B-Instruct at $T = 0.1$.

\begin{table}[h]
\centering
\caption{Dataset splits}
\small
\begin{tabular}{lrlc}
\toprule
Set & Size & Purpose & Disjoint from \\
\midrule
A & 5,000 & M1 diagnostic profiles & B, C$_1$ \\
B & 5,000 & Training (SFT pairs) & A, C$_1$ \\
C$_1$ & 3,000 & Evaluation & A, B \\
\bottomrule
\end{tabular}
\end{table}

C$_1$ domain distribution: Arts 847, Geography 581, History 956, Science 616.

\subsection{Training and evaluation pipeline}

Training used Llama-3-8B-Instruct with LoRA adapters in a custom PyTorch training loop. SFT with cross-entropy loss on response tokens only. Adapters were cast to fp32 to prevent NaN gradients under ROCm.

Evaluation used LoRA adapters merged with the base model, converted to f16 GGUF, and evaluated via llama-cpp-python with Vulkan backend on an AMD RX 7900 GRE at $T = 1.0$. All conditions were evaluated in f16 format.

Confidence was measured as NLP (mean token log-probability of the generated answer).

Meta-$d'$ was computed via Maniscalco and Lau (2012) MLE using the metadpy Python package (Legrand, 2022). NLP was binned into 8 quantile bins (2 $\times$ nRatings, nRatings $= 4$). Hautus log-linear correction ($+0.5$) was applied. The equal-variance assumption ($s = 1$) was used throughout.

\subsection{The quantisation format mismatch}

The M1 diagnostic profiles were computed using Q5\_K\_M GGUF quantisation via llama-cpp on the same GPU. The M2 evaluation used f16 GGUF because the LoRA merge pipeline could not produce Q5\_K\_M files. The pre-registration specified f16 to maintain internal consistency across adapted and baseline models.

This created a chain break. The prescription was derived from one measurement format and tested under another.

\subsection{Same-question format comparison}

To isolate the effect of quantisation from question-set differences, we evaluated the unadapted Llama-3-8B-Instruct on the identical C$_1$ evaluation set (3,000 questions) at both Q5\_K\_M and f16. This comparison was conducted after the confirmatory analysis revealed the null results and is reported as a post-hoc diagnostic analysis.

\subsection{Statistical tests}

All confirmatory tests used non-parametric bootstrap with 10,000 resamples at the question level within domain (seed $= 42$).

\textbf{H1} (Treatment effect): meta-$d'$(Cond 2) $-$ meta-$d'$(Cond 1) in Science, 95\% CI lower $> 0$.\\
\textbf{H2} (Non-degradation): TOST $|$meta-$d'$(Cond 2) $-$ meta-$d'$(Cond 1)$| < 0.17$ in History, Arts, Geography.\\
\textbf{H3} (Conditional advantage): meta-$d'$(Cond 2) $-$ meta-$d'$(Cond 3) in Science, 95\% CI lower $> 0$.\\
\textbf{H4} (Causal test): meta-$d'$(Cond 2) $-$ meta-$d'$(Cond 4) in Science, 95\% CI lower $> 0$.

\section{Results}

\subsection{M-ratio profiles restructure across formats}

Table \ref{tab:mratio} presents domain-level metacognitive profiles at Q5\_K\_M and f16 on the same 3,000 C$_1$ questions.

\begin{table}[h]
\centering
\caption{Domain-level metacognitive profiles under two quantisation formats. Same 3,000 C$_1$ questions, same model weights, same inference backend (llama-cpp, Vulkan, AMD RX 7900 GRE). Only the number format differs.}
\label{tab:mratio}
\small
\begin{tabular}{lcccccccc}
\toprule
& \multicolumn{4}{c}{Q5\_K\_M} & \multicolumn{4}{c}{f16} \\
\cmidrule(lr){2-5} \cmidrule(lr){6-9}
Domain & $d'$ & meta-$d'$ & M-ratio & Rank & $d'$ & meta-$d'$ & M-ratio & Rank \\
\midrule
Science & 0.365 & 0.493 & 1.352 & 1 & 0.461 & 0.663 & 1.436 & 2 \\
Geography & 0.410 & 0.497 & 1.210 & 2 & 0.648 & 0.518 & 0.798 & 3 \\
History & 0.675 & 0.415 & 0.615 & 3 & 0.722 & 0.339 & 0.470 & 4 \\
Arts & 0.891 & 0.540 & 0.606 & 4 & 0.559 & 0.862 & 1.542 & 1 \\
\bottomrule
\end{tabular}
\end{table}

\begin{figure}[h]
\centering
\includegraphics[width=0.95\textwidth]{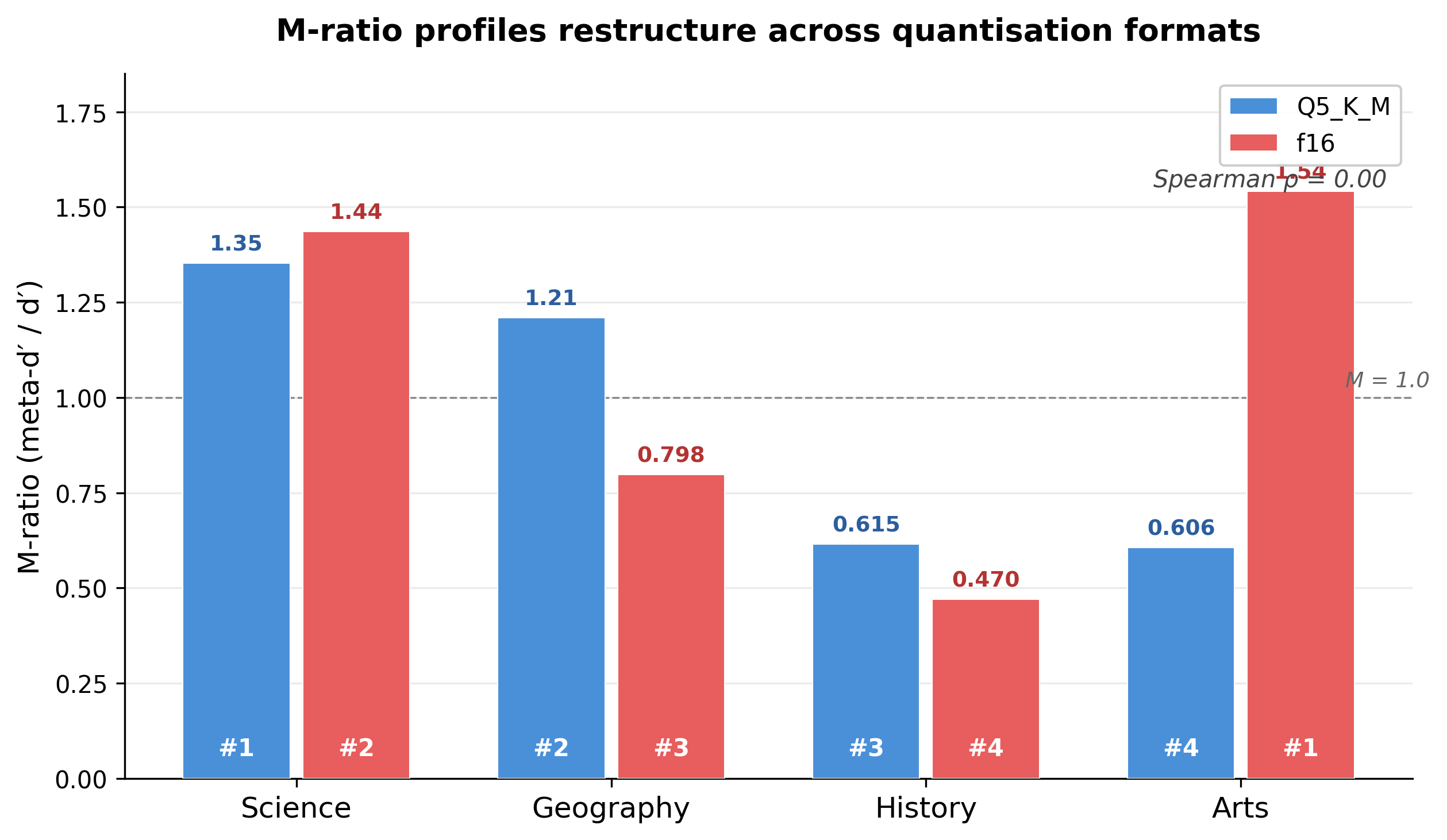}
\caption{M-ratio profiles by domain at Q5\_K\_M and f16 on the same 3,000 questions. Arts moves from rank 4 to rank 1. Geography moves from rank 2 to rank 3. Spearman $\rho = 0.00$.}
\label{fig:mratio}
\end{figure}

The M-ratio profile is not merely scaled. It is restructured. Arts moves from worst-monitored (0.606, rank 4) to best-monitored (1.542, rank 1). Geography moves from well-monitored (1.210, rank 2) to under-monitored (0.798, rank 3). The Spearman rank correlation between Q5\_K\_M and f16 M-ratio profiles is $\rho = 0.00$.

The $d'$ values shift non-proportionally across domains and formats. Arts $d'$ drops from 0.891 (Q5\_K\_M) to 0.559 (f16) while Science $d'$ increases from 0.365 to 0.461. These differential $d'$ shifts, combined with differential meta-$d'$ changes, produce the M-ratio restructuring. This is observed on the identical question set, eliminating item difficulty as a confound.

For comparison, the M1 diagnostic profiles (Set A, Q5\_K\_M) showed Arts as the worst-monitored domain at Q5\_K\_M in both samples, confirming cross-sample stability within format.

\subsection{AUROC$_2$ is format-stable}

To test whether the restructuring reflects a change in raw metacognitive discrimination or in the M-ratio normalisation specifically, we computed Type-2 AUROC per domain per format.

\begin{table}[h]
\centering
\caption{Type-2 AUROC per domain per format. Same 3,000 C$_1$ questions at both formats.}
\label{tab:auroc}
\small
\begin{tabular}{lcccc}
\toprule
Domain & Q5\_K\_M AUROC$_2$ & Q5\_K\_M Rank & f16 AUROC$_2$ & f16 Rank \\
\midrule
Arts & 0.710 & 1 & 0.680 & 1 \\
History & 0.669 & 2 & 0.672 & 2 \\
Geography & 0.629 & 3 & 0.668 & 3 \\
Science & 0.619 & 4 & 0.643 & 4 \\
\bottomrule
\end{tabular}
\end{table}

\begin{figure}[h]
\centering
\includegraphics[width=0.95\textwidth]{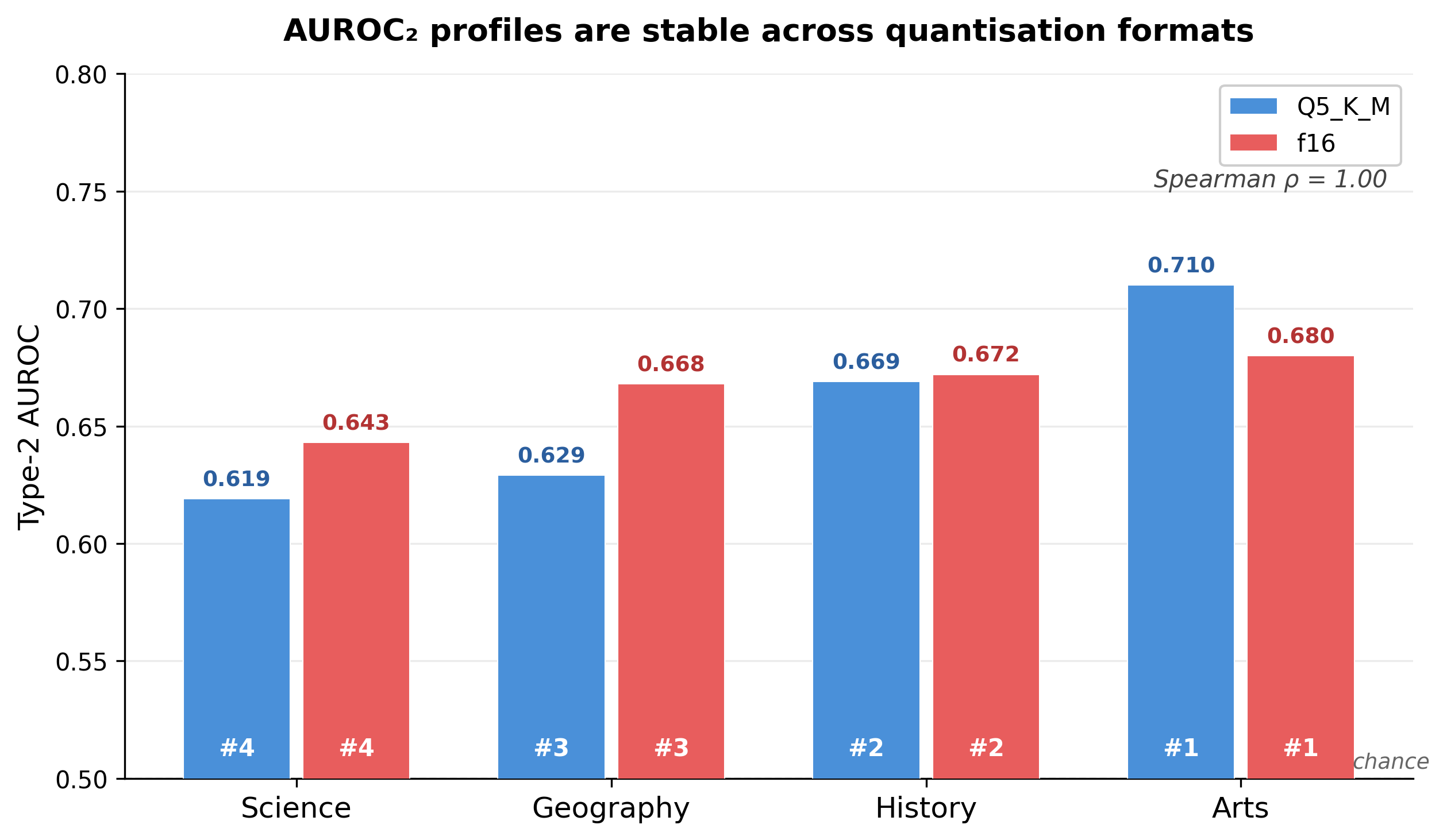}
\caption{AUROC$_2$ profiles by domain at Q5\_K\_M and f16 on the same 3,000 questions. Rank ordering is perfectly preserved. Spearman $\rho = 1.00$.}
\label{fig:auroc}
\end{figure}

The AUROC$_2$ rank ordering is perfectly preserved across formats ($\rho = 1.00$). Arts is the best-discriminated domain and Science the worst at both quantisation levels.

This dissociation is the central empirical result. M-ratio restructures ($\rho = 0.00$) while AUROC$_2$ is stable ($\rho = 1.00$). The raw Type-2 discrimination signal is format-invariant. What changes is the M-ratio, which divides meta-$d'$ by $d'$. Because quantisation shifts $d'$ non-proportionally across domains (Arts $d'$ drops by 37\% from Q5\_K\_M to f16, Science $d'$ increases by 26\%), the normalisation amplifies these differential shifts into an apparent profile restructuring.

\subsection{Confirmatory hypothesis tests}

\begin{table}[h]
\centering
\caption{Confirmatory results. 10,000 bootstrap resamples, seed $= 42$.}
\label{tab:confirmatory}
\small
\begin{tabular}{clllll}
\toprule
Hypothesis & Contrast & Domain & $\Delta$meta-$d'$ & 95\% CI & Result \\
\midrule
H1 & Cond 2 $-$ Cond 1 & Science & $-$0.118 & $[-0.539, 0.348]$ & Not supported \\
H3 & Cond 2 $-$ Cond 3 & Science & $-$0.041 & $[-0.460, 0.422]$ & Not supported \\
H4 & Cond 2 $-$ Cond 4 & Science & $+$0.138 & $[-0.223, 0.661]$ & Not supported \\
\bottomrule
\end{tabular}
\end{table}

\textbf{H2} (TOST, $\delta = 0.17$, 90\% CI):

\begin{table}[h]
\centering
\small
\begin{tabular}{lllc}
\toprule
Domain & $\Delta$meta-$d'$ & 90\% CI & Result \\
\midrule
History & $+$0.323 & $[-0.053, 0.543]$ & Not equivalent \\
Arts & $-$0.273 & $[-0.498, 0.121]$ & Not equivalent \\
Geography & $+$0.059 & $[-0.470, 0.324]$ & Not equivalent \\
\bottomrule
\end{tabular}
\end{table}

All four hypotheses are null. The wide H2 CIs reflect variability in meta-$d'$ estimation rather than evidence of degradation.

\subsection{The training did reshape confidence distributions}

Despite null meta-$d'$ effects, the SFT training successfully modified confidence behaviour.

\begin{table}[h]
\centering
\caption{NLP gap (mean NLP correct $-$ mean NLP incorrect) by condition and domain.}
\label{tab:nlpgap}
\small
\begin{tabular}{lrrrr}
\toprule
Condition & Science & Arts & History & Geography \\
\midrule
1 (Baseline) & 0.076 & 0.112 & 0.100 & 0.106 \\
2 (Conditional) & 0.152 & 0.136 & 0.147 & 0.156 \\
3 (Agnostic) & 0.123 & 0.152 & 0.128 & 0.143 \\
4 (Wrong-rx) & 0.133 & 0.162 & 0.141 & 0.115 \\
7 (Low LR) & 0.091 & 0.113 & 0.117 & 0.114 \\
\bottomrule
\end{tabular}
\end{table}

The Science NLP gap doubled under Condition 2 (0.076 to 0.152). The training did what it was designed to do. It amplified the separation between correct-trial and incorrect-trial NLP distributions. But amplifying confidence separation in a domain that is already well-monitored (M-ratio $= 1.436$ at f16) does not improve meta-$d'$.

Condition 7 (low learning rate) preserved accuracy almost exactly (0.676 vs 0.676 baseline) with modest NLP gap increases, confirming that learning rate controls the accuracy-confidence tradeoff.

\subsection{Full metrics table}

\begin{table}[h]
\centering
\caption{Per-condition, per-domain metrics.}
\label{tab:fullmetrics}
\small
\begin{tabular}{clrcccccr}
\toprule
Cond & Domain & N & Acc & $d'$ & meta-$d'$ & M-ratio & NLP gap \\
\midrule
1 & Arts & 847 & 0.647 & 0.559 & 0.862 & 1.542 & 0.112 \\
1 & Geography & 581 & 0.711 & 0.648 & 0.518 & 0.798 & 0.106 \\
1 & History & 956 & 0.668 & 0.722 & 0.339 & 0.470 & 0.100 \\
1 & Science & 616 & 0.696 & 0.461 & 0.663 & 1.436 & 0.076 \\
2 & Arts & 847 & 0.640 & 0.600 & 0.588 & 0.981 & 0.136 \\
2 & Geography & 581 & 0.673 & 0.689 & 0.577 & 0.837 & 0.156 \\
2 & History & 956 & 0.656 & 0.661 & 0.662 & 1.001 & 0.147 \\
2 & Science & 616 & 0.674 & 0.596 & 0.544 & 0.912 & 0.152 \\
3 & Arts & 847 & 0.638 & 0.594 & 0.746 & 1.255 & 0.152 \\
3 & Geography & 581 & 0.668 & 0.636 & 0.421 & 0.663 & 0.143 \\
3 & History & 956 & 0.641 & 0.517 & 0.684 & 1.323 & 0.128 \\
3 & Science & 616 & 0.679 & 0.447 & 0.585 & 1.309 & 0.123 \\
4 & Arts & 847 & 0.646 & 0.736 & 0.455 & 0.619 & 0.162 \\
4 & Geography & 581 & 0.685 & 0.474 & 0.410 & 0.866 & 0.115 \\
4 & History & 956 & 0.660 & 0.652 & 0.578 & 0.887 & 0.141 \\
4 & Science & 616 & 0.672 & 0.616 & 0.406 & 0.659 & 0.133 \\
7 & Arts & 847 & 0.655 & 0.581 & 0.804 & 1.384 & 0.113 \\
7 & Geography & 581 & 0.697 & 0.667 & 0.462 & 0.693 & 0.114 \\
7 & History & 956 & 0.675 & 0.656 & 0.573 & 0.873 & 0.117 \\
7 & Science & 616 & 0.687 & 0.384 & 0.626 & 1.631 & 0.091 \\
\bottomrule
\end{tabular}
\end{table}

\section{Discussion}

\subsection{What restructures and what does not}

The central empirical result is a dissociation. M-ratio profiles restructure across quantisation formats ($\rho = 0.00$). AUROC$_2$ profiles are perfectly stable ($\rho = 1.00$). The same model on the same questions shows identical Type-2 discrimination ordering at Q5\_K\_M and f16. Arts best, Science worst. But the M-ratio ordering is completely different.

This dissociation localises the format-dependence precisely. AUROC$_2$ is a rank-based, non-parametric measure of whether confidence discriminates correct from incorrect responses. It is insensitive to $d'$. M-ratio divides meta-$d'$ by $d'$, normalising metacognitive sensitivity for task difficulty. Quantisation shifts $d'$ non-proportionally across domains. Arts $d'$ drops 37\%. Science $d'$ increases 26\%. These differential shifts propagate through the normalisation to produce an apparent profile restructuring that does not exist in the raw discrimination signal.

We propose the term \textit{metacognitive geometry} for the domain-level structure of M-ratio profiles. The geometry is format-dependent because $d'$ is format-dependent and M-ratio inherits that dependency through division. This is not a change in the model's underlying capacity to discriminate correctness from confidence. It is a change in the normalised efficiency metric.

This goes beyond the findings of Proskurina et al. (2024), who showed that quantisation decreases confidence on true labels and disproportionately affects low-confidence samples. Their result describes a level shift. Our result describes a normalisation-mediated structural shift. The distinction matters operationally. A level shift can be corrected by recalibration (as Liu et al., 2025 demonstrate). A normalisation-mediated structural shift can be avoided by using format-stable metrics like AUROC$_2$ instead of M-ratio for diagnostic purposes.

This parallels findings at two other levels of analysis. Cacioli (2026d) showed that 12 of 20 frontier models produce inconsistent validity profiles across retrospective and prospective probe formats. Cacioli (2026e) demonstrated that transformers reproduce the geometry of magnitude representations but not the fine-grained noise profiles. The raw Type-2 discrimination ordering (AUROC$_2$) is format-stable. The normalised efficiency metric (M-ratio) is not.

\subsection{Implications for diagnostic-guided systems}

The prescribe-don't-average framework remains theoretically sound. Domain-conditional interventions should outperform domain-agnostic ones when the diagnosis is valid. The present study reveals that the choice of diagnostic metric determines format-sensitivity.

Had M1 diagnosed the weakest domain via AUROC$_2$ rather than M-ratio, the prescription would have been Science at both formats and the format mismatch would not have arisen.

The practical recommendation is twofold. First, diagnostic-guided systems should prefer format-stable metrics (AUROC$_2$, NLP gap) over format-sensitive ones (M-ratio) when diagnosis and deployment may use different inference formats. Second, when M-ratio is used for its theoretical grounding in the Type-2 SDT framework, format consistency between diagnosis and evaluation is a measurement requirement.

\subsection{The NLP gap dissociation}

An important subsidiary finding is that the NLP gap can increase without improving meta-$d'$. Condition 2 doubled the Science NLP gap (0.076 to 0.152) but produced $\Delta$meta-$d' = -0.118$. This dissociation arises because meta-$d'$ depends not just on the mean separation between correct and incorrect NLP distributions but on their overlap structure. A mean shift that does not improve the overlap structure does not improve metacognitive sensitivity.

This has practical implications for calibration training. Optimising for NLP gap or similar first-moment statistics is not sufficient for improving metacognitive discrimination.

\subsection{Condition 7 and the accuracy-metacognition tradeoff}

Condition 7 (LR $=$ 5e-6, half the standard rate) preserved accuracy almost perfectly (0.676 vs 0.676 baseline) while producing the highest Science meta-$d'$ (0.626) and M-ratio (1.631) of any condition. This was exploratory and not pre-registered as confirmatory. The extremely high M-ratio should be interpreted with caution given the low $d'$ (0.384), where M-ratio becomes unstable (Fleming \& Lau, 2014).

\subsection{Honest reporting of a null result}

This study was pre-registered with all hypotheses, analyses, and decision rules specified before training. The results are null. We report them as specified because pre-registered null results are informative. The null was caused by a specific, identifiable, and novel measurement property rather than by a failure of the theoretical framework. Post-hoc modification of the analysis to accommodate the format mismatch would violate the pre-registration.

\subsection{Limitations}

The same-question comparison (Table \ref{tab:mratio}) eliminates the question-set confound for the format comparison. However, the confirmatory training intervention was conducted entirely at f16. Whether the intervention would succeed if both diagnosis and evaluation used Q5\_K\_M remains untested.

The study examines one model at two quantisation levels. Whether the metacognitive geometry restructuring generalises across models, model scales, and quantisation methods (GPTQ, AWQ, INT8) is unknown. The pre-registered replication on Mistral-7B was not conducted due to the null primary result.

Several conditions produce $d'$ values below 0.5 (Q5\_K\_M Science $d' = 0.365$, f16 Science $d' = 0.461$), where M-ratio becomes unstable (Fleming \& Lau, 2014). The profile restructuring should be interpreted alongside the absolute $d'$ values, not from M-ratio alone. Future work should report AUROC$_2$ as a complementary metric that avoids the $d'$-denominator instability.

The mechanism by which quantisation produces domain-differential $d'$ compression is unknown. Whether it relates to domain-specific token distributions, attention head sensitivity, or weight magnitude distributions remains to be investigated.

\section{Conclusion}

We set out to improve domain-specific metacognition through targeted training. Instead we discovered that the diagnostic metric we used is format-dependent. Quantisation restructures M-ratio profiles across domains because it shifts $d'$ non-proportionally and M-ratio inherits those shifts through division. The underlying discrimination ordering (AUROC$_2$) is stable.

The practical implication is that M-ratio-based diagnostics require format consistency between diagnosis and deployment. AUROC$_2$-based diagnostics do not. For any system that diagnoses, acts on, or deploys domain-level confidence properties, the choice of metric determines whether the diagnosis transfers across inference formats.

\section*{Open Science}

Pre-registration 1 (pilot design and decision gate) is posted at \url{https://osf.io/8c6kp/}. Pre-registration 2 (full confirmatory grid) is posted at \url{https://osf.io/juhcr/}. All training scripts, evaluation code, trial-level data, and analysis outputs are publicly available at \url{https://github.com/synthiumjp/sdt-calibration}. The confirmatory analysis is fully reproducible (seed $= 42$, 10,000 bootstrap resamples).

\section*{Generative AI Disclosure}

Claude (Anthropic) was used for pipeline development, code generation, and manuscript preparation. All scientific decisions, interpretations, and the identification of the quantisation confound were made by the author.

\section*{References}

\small

Cacioli, J.-P. (2026a). Domain-specific metacognitive efficiency in large language models: A Type-2 signal detection theory analysis. \textit{arXiv:2603.25112}.

Cacioli, J.-P. (2026d). Before you interpret the profile: Validity scaling for LLM metacognitive self-report. \textit{Manuscript in preparation}.

Cacioli, J.-P. (2026e). Scalar variability in transformer language models. \textit{Manuscript in preparation}.

Dettmers, T., Lewis, M., Belkada, Y., \& Zettlemoyer, L. (2022). GPT3.int8(): 8-bit matrix multiplication for transformers at scale. \textit{NeurIPS}.

Fleming, S. M., \& Lau, H. C. (2014). How to measure metacognition. \textit{Frontiers in Human Neuroscience}, 8, 443.

Frantar, E., Ashkboos, S., Hoefler, T., \& Alistarh, D. (2023). GPTQ: Accurate post-training quantization for generative pre-trained transformers. \textit{ICLR}.

Guo, C., Pleiss, G., Sun, Y., \& Weinberger, K. Q. (2017). On calibration of modern neural networks. \textit{ICML}.

Legrand, N. (2022). metadpy: Metacognitive efficiency modelling in Python. \url{https://github.com/LegrandNico/metadpy}.

Liu, S., Li, Z., Liu, X., Zhan, R., Wong, D., Chao, L., \& Zhang, M. (2025). Quantized can still be calibrated. \textit{ACL 2025}.

Maniscalco, B., \& Lau, H. (2012). A signal detection theoretic approach for estimating metacognitive sensitivity from confidence ratings. \textit{Consciousness and Cognition}, 21(1), 422--430.

Parikh, N., Sai, A., Shivaswamy, P., Panchal, K., \& Lan, A. (2026). CATTO: Balancing preferences and confidence in language models. \textit{arXiv:2601.23096}.

Proskurina, I., Brun, L., Metzler, G., \& Velcin, J. (2024). When quantization affects confidence of large language models? \textit{Findings of NAACL 2024}, 1918--1928.

Singh, M., Durrani, N., Dalvi, F., \& Sajjad, H. (2025). Interpreting the effects of quantization on LLMs. \textit{arXiv:2508.16785}.

Steyvers, M., \& Peters, M. A. K. (2025). Metacognition and uncertainty communication in humans and large language models. \textit{arXiv:2504.14045}.

Steyvers, M., Belem, C., \& Smyth, P. (2025). Improving metacognition and uncertainty communication in language models. \textit{arXiv:2510.05126}.

\end{document}